%% file: aaai2026.tex
\documentclass[letterpaper]{article} 
\usepackage{aaai2026}  
\nocopyright
\usepackage{times}  
\usepackage{helvet}  
\usepackage{courier}  
\usepackage[hyphens]{url}  
\usepackage{graphicx} 
\def\UrlFont{\rm}  
\usepackage{natbib}  
\usepackage{caption} 
\usepackage{algorithm}
\usepackage{algorithmic}

\usepackage{newfloat}
\usepackage{listings}
\DeclareCaptionStyle{ruled}{labelfont=normalfont,labelsep=colon,strut=off} 
\floatstyle{ruled}
\newfloat{listing}{tb}{lst}{}
\floatname{listing}{Listing}
\input{math_commands}

\usepackage{multirow}
\usepackage{booktabs}
\usepackage{array}
\usepackage{soul}
\usepackage[table]{xcolor}
\usepackage{subfig}

\definecolor{LightCyan}{RGB}{232,241,255}
\definecolor{LightRed}{RGB}{255,235,235}
\definecolor{LightPink}{RGB}{255,235,255}
\definecolor{LightGreen}{RGB}{218,255,234}
\definecolor{LightYellow}{RGB}{255,255,235}
\definecolor{Gray}{RGB}{242,242,242}
\definecolor{Red}{RGB}{253, 239, 242}
\definecolor{Yellow}{RGB}{255, 255, 204}
\definecolor{Pink}{RGB}{255, 243, 254}
\definecolor{LightGray}{RGB}{249, 249, 249}
\definecolor{Green}{RGB}{230, 255, 241}
\definecolor{Blue1}{RGB}{239, 248, 253}
\definecolor{Blue2}{RGB}{218, 232, 245}
\definecolor{Blue3}{RGB}{136, 190, 220}
\definecolor{Blue4}{RGB}{83, 157, 204}
\definecolor{Blue5}{RGB}{42, 122, 185}
\definecolor{Blue6}{RGB}{11, 85, 159}
\definecolor{GreenCheck}{RGB}{0, 102, 51}
\definecolor{LightBack}{RGB}{247,249,251}
\definecolor{DarkRed}{HTML}{C00000}
\definecolor{DarkGreen}{HTML}{3B7D23}

\title{From Personal to Collective:\\ On the Role of Local and Global Memory in LLM Personalization}

\author{
    Zehong Wang$^{1}$,
    Junlin Wu$^{3}$,
    Zhaoxuan Tan$^{1}$,
    Bolian Li$^{4}$,
    Xianrui Zhong$^{5}$,\\
    Zheli Liu$^{2}$,
    Qingkai Zeng$^{2}$\thanks{{ }{ }Corresponding author.}
}
\affiliations{
    \textsuperscript{\rm 1}Department of Computer Science and Engineering, University of Notre Dame\\
    \textsuperscript{\rm 2}College of Computer Science, Nankai University, \\
    \textsuperscript{\rm 3}Washington University in St. Louis
    \textsuperscript{\rm 4}Purdue University
    \textsuperscript{\rm 5}UIUC \\
    zwang@nd.edu, junlin.wu@wustl.edu, qzengnkcs@gmail.com

}

\usepackage{bibentry}

\begin{document}

\maketitle

\begin{abstract}
    Large language model (LLM) personalization aims to tailor model behavior to individual users based on their historical interactions. However, its effectiveness is often hindered by two key challenges: the \textit{cold-start problem}, where users with limited history provide insufficient context for accurate personalization, and the \textit{biasing problem}, where users with abundant but skewed history cause the model to overfit to narrow preferences. We identify both issues as symptoms of a common underlying limitation, i.e., the inability to model collective knowledge across users. To address this, we propose a local-global memory framework (LoGo) that combines the personalized local memory with a collective global memory that captures shared interests across the population. To reconcile discrepancies between these two memory sources, we introduce a mediator module designed to resolve conflicts between local and global signals. Extensive experiments on multiple benchmarks demonstrate that LoGo consistently improves personalization quality by both warming up cold-start users and mitigating biased predictions. These results highlight the importance of incorporating collective knowledge to enhance LLM personalization.

\end{abstract}

\section{Introduction}

Personalization \citep{chen2024persona,raji2024commerce,chen2024large,deldjoo2024review} plays a central role in modern AI systems, enabling models to adapt their behavior to the unique preferences, goals, and interaction histories of individual users \citep{tan2024oppu,gupta2024rag}. Across domains such as e-commerce \citep{ag2024personalized}, entertainment \citep{trifts2019enhancing}, education \citep{tetzlaff2021developing}, and productivity \citep{kim2019understanding}, personalization enhances usability, increases engagement, and delivers more relevant and meaningful experiences. At its core, personalization leverages past user interactions to inform future responses, improving decision quality, and aligning system outputs with user intent.

Thanks to their emergent capabilities and massive parameter scales, Large Language Models (LLM) have reshaped the landscape of natural language processing. Recent LLM personalization methods model user preferences by treating historical interactions as textual sequences, enabling LLMs to retrieve~\cite{salemi2023lamp,salemi2024optimization}, summarize~\cite{richardson2023integrating}, and condition on user-specific signals~\cite{zhuang2024hydra}. One notable approach, One PEFT Per User (OPPU)~\cite{tan2024oppu}, assigns each user a dedicated, parameter-efficient module, breaking from the one-size-fits-all paradigm to enable fine-grained adaptation. 

\begin{figure}[!t]
    \centering
    \includegraphics[width=\linewidth]{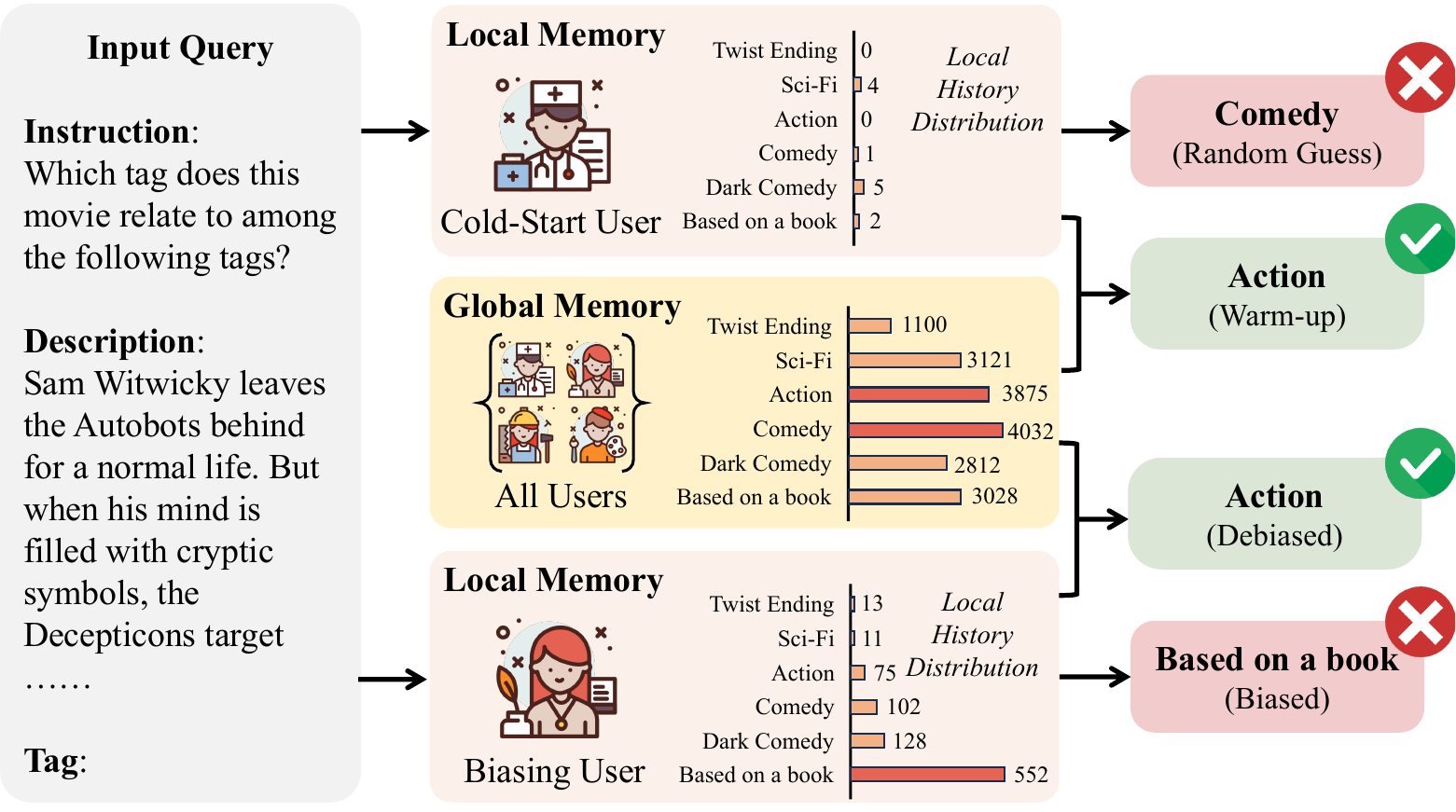}
    \caption{Global memory improves personalization by addressing data sparsity and bias. For cold-start users, it compensates for limited local history with population-wide knowledge. For users with skewed histories, it helps correct biased predictions by aligning to boarder user behaviors.}
    \label{fig:motivation}
    \vspace{-0.2in}
\end{figure}

Despite their effectiveness, these user-centric methods, which rely solely on individual histories for personalization, face two key challenges: the cold-start problem for users with sparse data, and prediction bias for users with abundant but skewed histories.
We posit that both challenges arise from a single root: the inability to model the collective knowledge shared across users. Incorporating such collective knowledge enables better generalization in low-data scenarios and mitigates overfitting in high-data settings, aligning to the social norms and commonsense priors that provide principled guidance when user data is limited or overly specific. 
As illustrated in Figure \ref{fig:motivation}, collective knowledge (global memory) complement individual traces (local memory), leading to more accurate and less biased predictions. 
This leads us to ask:
\begin{center}
    \textit{What is the role of collective knowledge in enhancing LLM personalization?}
\end{center}

To answer this question, we explore modeling shared knowledge across users for LLM personalization, a non-trivial task that is considerably more complex than modeling individual user interests alone. This setting introduces two key challenges: (1)~\textit{Evolution}: Global knowledge is inherently dynamic, continuously shaped by the shifting interests and behaviors of users over time \citep{wu2017modeling}. Accurately capturing this temporal evolution is crucial for reflecting current trends and avoiding outdated generalizations. (2)~\textit{Conflict}: Individual preferences may diverge from population-level patterns \citep{zheng2021disentangling}, creating tension between personalization and generalization. These two challenges reflect insights from social science, where social norms emerge and evolve through collective behavior but often contradict personal viewpoints \citep{hechter2001social}. Several recent studies have explored integrating shared knowledge into LLM personalization, but often tackle only one aspect of the problem. For example, PER-PCS~\citep{tan2024personalized} enhances user representations by aggregating information from related users, but failing to reconcile conflicts between collective patterns and user-specific preferences. HYDRA~\citep{zhuang2024hydra} employs a frozen shared backbone to inject collective knowledge and user-specific heads for personalization, but failing to model evolving community trends. 

To this end, we propose LoGo, a \textbf{Lo}cal–\textbf{G}l\textbf{o}bal memory framework that explicitly models both personalized and collective interests for LLM personalization. LoGo is a flexible framework that can be implemented in a parametric way for white-box LLMs (main paper) and a non-parametric way for black-box settings (appendix). 
In particular, the \textit{local memory} captures user-specific behavior, while the \textit{global memory} encodes collective knowledge shared across users. This dual-memory architecture enables the system to generalize in cold-start scenarios and mitigate overfitting in biased user histories. To reconcile discrepancies between local and global signals, we introduce a \textit{mediator} module to balance their contributions. Additionally, we extend the framework with a \textit{community-level memory} component to capture shared knowledge at a finer granularity by grouping similar users. 
Experiments on five personalization benchmarks show that global memory consistently improves performance, boosting prediction accuracy for cold-start users and enhancing output diversity for highly active ones.

\begin{figure*}[!ht]
    \centering
    \includegraphics[width=\linewidth]{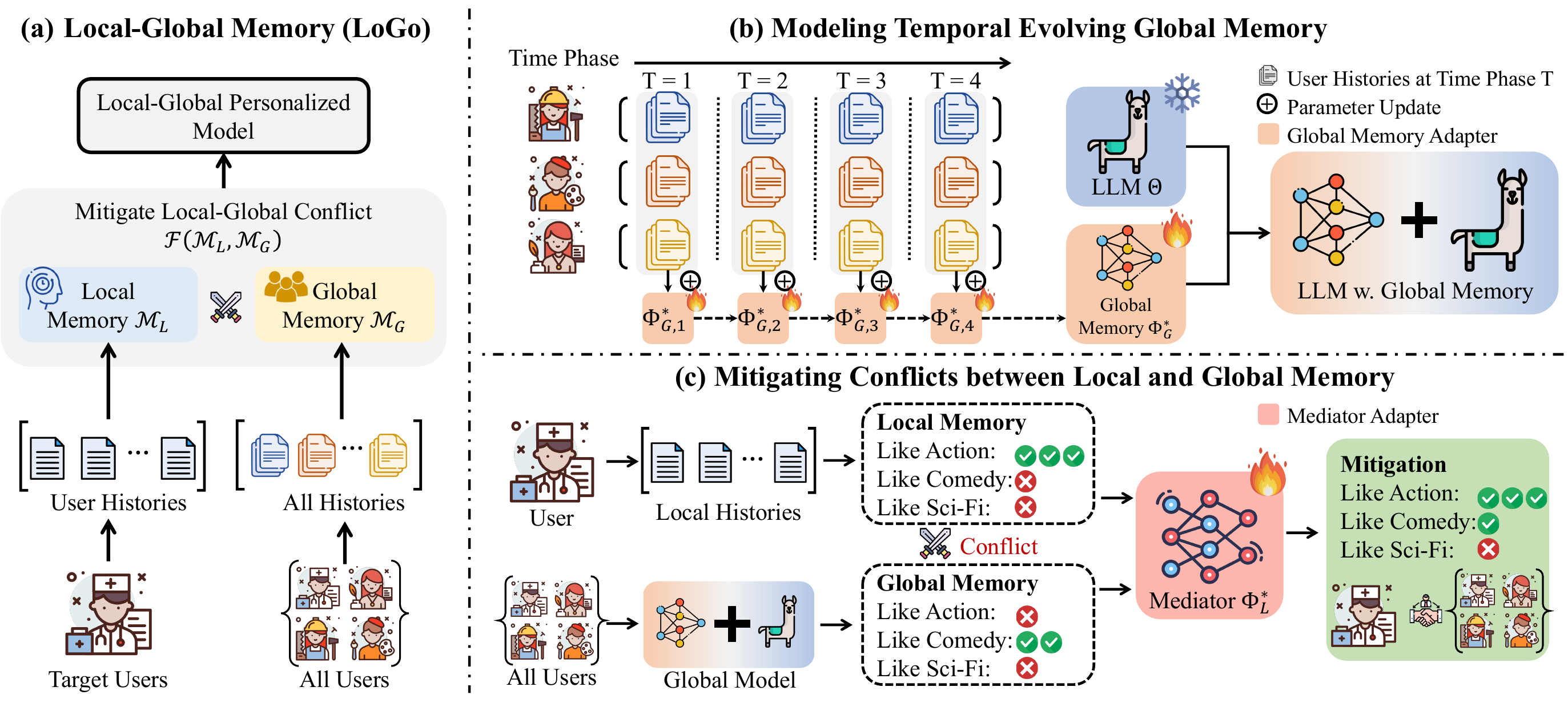}
    \caption{Overview of the local-global memory framework (LoGo) for LLM personalization. (a) The model integrates local memory, which captures user-specific history, with global memory that encodes shared knowledge across users. (b) Global memory is updated over time through temporal phases to reflect evolving user interests. (c) A mediator module resolves conflicts between local and global signals, enabling balanced personalization and generalization.}
    \label{fig:framework}
    \vspace{-0.1in}
\end{figure*}

\input{method}

\begin{figure*}[!t]
    \centering
    \subfloat[Performance when limiting the maximum number of histories per user.]{\includegraphics[width=\linewidth]{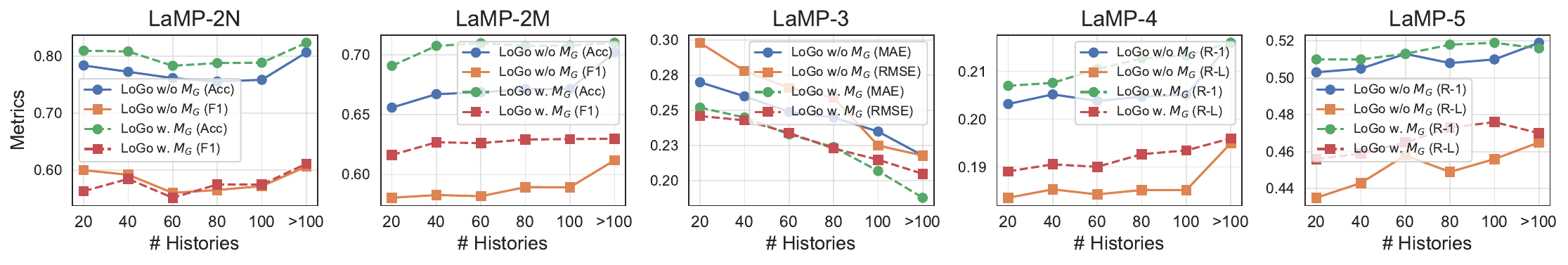}\label{fig:scaling-up history}}\\
    \subfloat[Performance when varying the number of users used to construct the global memory.]{\includegraphics[width=\linewidth]{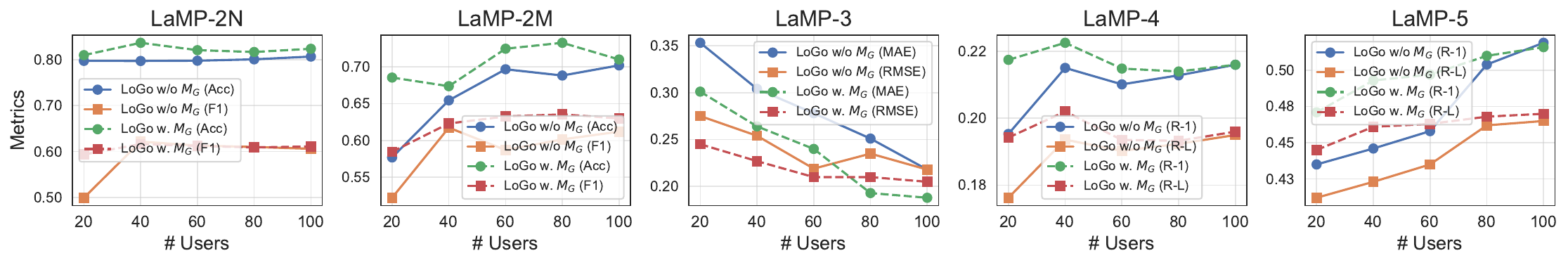}\label{fig:scaling-up user}}
    \caption{
    Generalization performance under varying data availability. The model demonstrates strong generalization even with limited histories per user (a), and maintains robust performance with fewer users contributing to the global memory (b).
    }
    \label{fig:scaling-up}
\end{figure*}

\section{Experiments}

We evaluate our LoGo using the LaMP benchmark~\citep{salemi2023lamp}, covering classification, regression, and text generation tasks, using task-specific metrics: Accuracy/F1 for classification, MAE/RMSE for regression, and  ROUGE-1/L for text generation. We consider both white-box and black-box settings using LLaMA 3.1 8B \citep{dubey2024llama} and Claude 3.7 as the respective backbone models. Following~\citet{tan2024oppu}, we evaluate performance on the top 100 most active users, identified based on the length of their interaction histories,  while the remaining users are used to train the base LLM. For retrieval-based methods, we use BM25 \citep{robertson2009probabilistic} with one retrieved item by default for consistency.

\input{table/cold-start}

\subsection{Main Results}

To demonstrate the effectiveness of LoGo, we conduct experiments across a broad spectrum of baselines, including: the base model and its variant with randomly retrieved items; OPPU \citep{tan2024oppu}, as well as retrieval-augmented variants such as RAG \citep{gupta2024rag} and PAG \citep{richardson2023integrating}; PER-PCS~\citep{tan2024personalized} and HYDRA~\citep{zhuang2024hydra} that leverage shared knowledge to improve the generalization. 
For our LoGo, we consider two settings: the \textit{base version} and the \textit{community version}. In the base version, we evaluate three variants: the base model, a RAG version, and a PAG version. In the community version, we use the PAG variant as the base and vary the number of clusters to assess performance under different community sizes. The experimental results are shown in Table~\ref{tab:parametric}.

\input{table/biasing}

\begin{figure*}[!t]
    \centering
    \subfloat[Cosine similarity between global memories across different time phases.]{\includegraphics[width=\linewidth]{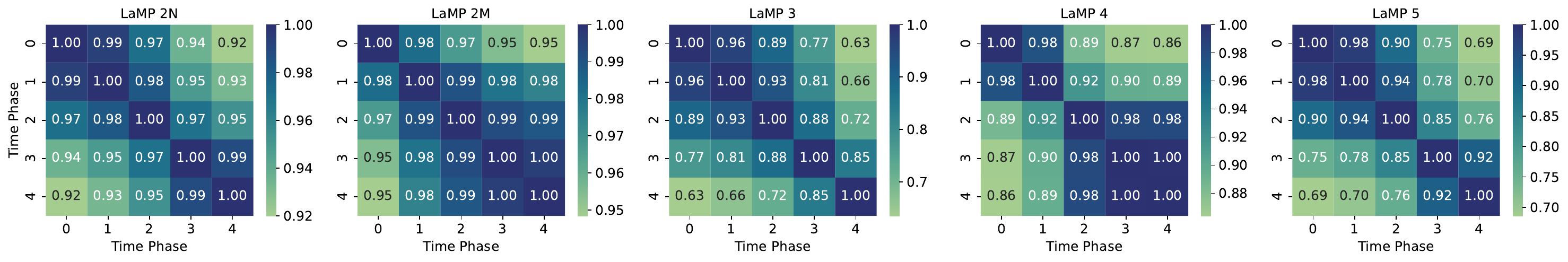}\label{fig:global evolving}}
    
    \subfloat[Model performance on LaMP-2M.]{\includegraphics[width=0.49\linewidth]{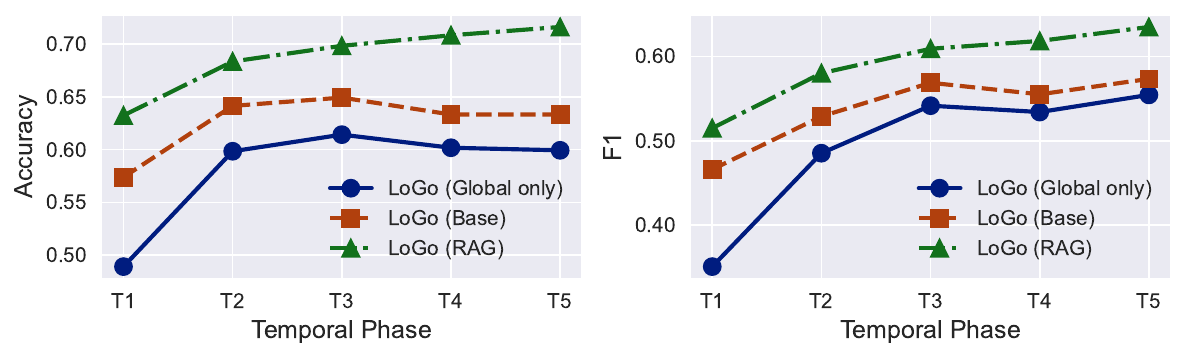}\label{fig:shift movie tagging}}
    \subfloat[Model performance on LaMP-4.]{\includegraphics[width=0.49\linewidth]{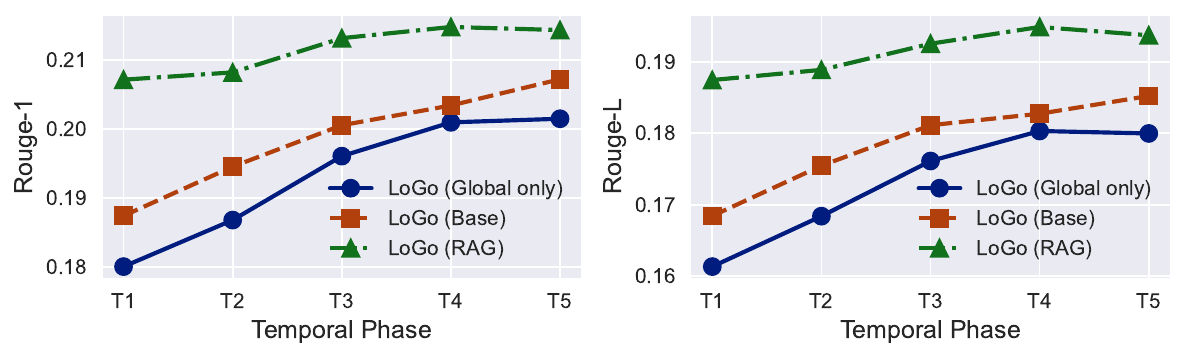}\label{fig:shift news headline}}
    \caption{Performance under temporal distribution shift. (a) Cosine similarity between global memories across different time phases, illustrating temporal consistency or drift. (b) Performance on LaMP-2M and (c) LaMP-4 across temporal phases, demonstrating that augmenting global memory with local memory consistently improves performance over time.}
    \label{fig:distribution shift}
\end{figure*}

\noindent\textbf{Effectiveness of LoGo.} 
 Overall, personalized methods consistently outperform non-personalized ones, demonstrating the effectiveness of personalization. Our LoGo further improves performance over existing personalization baselines by leveraging global memory that captures shared knowledge across users as well as mitigates the conflict between the personalized and collective interests. 

\noindent\textbf{Effect of Community Clusters.} 
To evaluate the impact of community-aware global memory, we vary the number of user clusters $c \in \{5, 10, 20\}$. Performance improves significantly when increasing $c$ from 1 to 5 or 10, suggesting that finer-grained user grouping helps capture more meaningful shared knowledge. However, setting $c=20$ leads to diminishing or inconsistent gains, likely due to over-fragmentation and reduced generalization. These results indicate that incorporating community-level global memory enhances performance by effectively modeling fine-grained shared knowledge, but an optimal cluster size is critical to balance specificity and generalizability.

\subsection{Cold-Start and Debiasing Analysis}

\noindent\textbf{Warming Up Cold-Start Users.}  
Table~\ref{tab:cold-start} presents the evaluation results for inactive (cold-start) users, those in the bottom 25\% based on the number of historical interactions, across both the LaMP-2M (classification) and LaMP-4 (text generation) tasks. We evaluate two variants of LoGo: the base version and the RAG-enhanced version. The results show that incorporating global memory $\mathcal{M}_G$ consistently enhances model performance, with improvements of up to 18.05\%. These findings highlight the effectiveness of leveraging collective knowledge to improve cold-start performance, enabling LoGo to overcome sparse user-specific signals by drawing on shared behavioral patterns.

\noindent\textbf{Mitigating Bias in High-Activity Users.}  
Table~\ref{tab:biasing} reports the experimental results on highly active users, defined as the top 25\% of users with the richest historical interactions. This subgroup is particularly prone to prediction bias, often caused by overfitting to individualized behavior patterns. 
We evaluate both prediction performance and diversity ratio to demonstrate the debiasing capability of our LoGo model. 
In particular, we use the normalized entropy of each user’s prediction distribution as a surrogate measure of diversity, with a detailed definition provided in the appendix.
The results demonstrate that using only local memory can lead to excessive personalization, limiting generalization and reducing output diversity. In contrast, augmenting the model with global memory $\mathcal{M}_G$ consistently improves both predictive performance and diversity. This suggests that global memory acts as a form of population-level regularization, helping the model balance between personalization and general trends, ultimately mitigating user-specific biases.

\input{table/ablation.tex}

\subsection{Generalization Analysis}

The effectiveness of global memory in mitigating cold-start issues and reducing bias for high-activity users suggests that it captures generalizable patterns beyond individual user histories. To further evaluate this generalization capability, we examine performance under two constrained settings: (1) limiting the number of historical interactions per user, and (2) restricting the number of users contributing to global memory. The results are presented in Figure~\ref{fig:scaling-up}.

\noindent\textbf{Limited History Setting.}  
We evaluate model performance when restricting each user to a maximum of $n$ past interactions. Unlike the cold-start scenario, this setting introduces a temporal distribution shift, as earlier interactions may not align well with the current query context. As shown in Figure~\ref{fig:scaling-up history}, global memory significantly enhances performance when history is sparse (e.g., $n=20$ or $n=40$), effectively compensating for the lack of local signals. These results underscore the generalization capability of global knowledge in underdetermined personalization settings.

\noindent\textbf{Limited User Setting.}  
We assess model robustness when limiting the number of users contributing to global memory construction. As shown in Figure~\ref{fig:scaling-up user}, the performance gains from incorporating global memory are most pronounced when user coverage is low. This suggests that the learned global knowledge remains transferable even with limited population diversity. However, as the number of contributing users increases, the performance improvements begin to plateau, indicating that overly broad aggregation may introduce noise or dilute informative signals. These findings highlight the importance of designing fine-grained strategies for global modeling.

\subsection{Temporal Distribution Shift Analysis}

The global memory captures temporal information, which introduces temporal distribution shift. 
We analyze the impact of such distribution shift on model performance.

\noindent\textbf{Evolution of Global Memory.}  
To evaluate the stability of global knowledge over time, we compute the pairwise cosine similarity between global memory parameters $\Phi_{G, t}^*$ across different temporal phases. High similarity indicates slow evolution and stable preferences, whereas low similarity reflects rapidly shifting global trends. As shown in Figure~\ref{fig:global evolving}, datasets such as LaMP-2N, LaMP-2M, and LaMP-4 exhibit stable dynamics, with similarity consistently above 0.85 across phases. In contrast, LaMP-3 and LaMP-5 show more rapid evolution, particularly in later phases, suggesting greater temporal drift in user behavior.

\noindent\textbf{Robustness to Temporal Shift.}  
We further assess model robustness by evaluating performance across temporal phases (T1–T5). As shown in Figures~\ref{fig:shift movie tagging} and \ref{fig:shift news headline}, models relying solely on global memory exhibit performance degradation in earlier phases due to distributional mismatch. In contrast, combining global and local memory yields consistently strong results, even under severe temporal shifts.

\subsection{Hyper-parameter Analysis}

We perform hyper-parameter analysis using LoGo-PAG as the backbone, where the results are shown in Table~\ref{tab:ablation}. 

\noindent\textbf{Effect of Temporal Resolution.}  
We examine the impact of varying the number of time splits $T \in \{1, 5, 10\}$ used to update the global memory. Increasing $T$ consistently improves performance by capturing more fine-grained temporal dynamics. Notably, setting $T=1$ leads to a substantial performance drop, underscoring the importance of modeling evolving user preferences over time.

\noindent\textbf{Effect of Retrieved Local Memory.}  
Finally, we examine the impact of varying the number of retrieved local history items, $k \in \{1, 2, 4\}$. We observe consistent performance improvements as $k$ increases, indicating that incorporating more relevant local context enhances the model’s ability to reconcile global and personal preferences. These results underscore the value of retrieval-based local memory in supporting effective personalization.

\section{Related Work}

Existing methods for LLM personalization can be broadly categorized into non-parametric and parametric approaches. Non-parametric methods, primarily based on in-context learning, personalize outputs by conditioning on user-specific information without modifying model parameters. Prior works~\cite{zhiyuli2023bookgpt, kang2023llms, wang2023learning} have demonstrated that incorporating user behavior histories as few-shot examples enables LLMs to generate personalized responses across diverse tasks. To enhance the relevance of selected examples, retrieval-augmented prompting~\cite{salemi2023lamp, salemi2024optimization, mysore2023pearl, li2023teach} has proven effective in identifying pertinent records from a user's behavioral history. Beyond simple retrieval, recent efforts have proposed profile-augmented prompting~\cite{richardson2023integrating}, which summarizes user preferences and behavioral patterns in natural language to enrich the input. Others have explored hierarchical personalized retrieval systems~\cite{sun2024persona} to better structure user-specific context. In contrast, parametric methods adapt model weights to explicitly capture individual user preferences. OPPU~\cite{tan2024oppu} fine-tunes per-user models using LoRA~\cite{hu2021lora}. Other recent work explores model merging for alignment~\cite{jang2023personalized}, personalized RLHF (pRLHF)~\cite{park2024rlhf, li2024personalized}, and user-specific reward modeling~\cite{cheng2023deserves}. While these methods focus on modeling personalization at the individual level, they generally do not account for knowledge shared across users. Some recent efforts attempt to incorporate shared knowledge, such as PER-PCS~\cite{tan2024personalized}, which uses a shared LoRA pool, and HYDRA~\cite{zhuang2024hydra}, which adopts a shared backbone model with user-specific heads. However, these methods fall short of jointly modeling the temporal evolution of shared knowledge and resolving conflicts between individual and collective preferences. Our framework is designed to address these limitations.

\section{Conclusion}

In this paper, we investigate LLM personalization from a new perspective—bridging individual preferences with collective user knowledge. We introduce a local-global memory framework (LoGo) that combines user-specific local memory with a global memory capturing shared patterns across users. To reconcile discrepancies between these two sources, we propose a mediator that adaptively balances personalization and generalization. Extensive experiments across multiple benchmarks show that our framework effectively enhances personalization quality, particularly by addressing cold-start limitations and mitigating user-specific bias.

\bibliography{aaai2026}


\newpage
\appendix

\section{The Diversity Measurement}

To evaluate the biasing ratio of user predictions, we define a diversity measurement where a large diversity indicates low biasing. To this end, we define an entropy-based method to measure the diversity of the prediction results. Take classification tasks as an example, each user \( u \) is associated with a set of predicted labels from a classifier:
\[
L_u = \{\ell_1, \ell_2, \ldots, \ell_m\}, \quad \ell_i \in \{1, 2, \ldots, n\}
\]
From this, we define a discrete empirical distribution \( P_u = \{p_{u,1}, p_{u,2}, \ldots, p_{u,n} \} \), where:
\[
p_{u,i} = \frac{\text{count of label } i \text{ in } L_u}{m}
\]
That is, \( p_{u,i} \) is the relative frequency of label \( i \) among the \( m \) predictions for user \( u \), and \( \sum_{i=1}^{n} p_{u,i} = 1 \).
Given a distribution $P_u$, we define the diversity using normalized entropy as: 
\[
Div(P_u) = \frac{H(P_u)}{H_{\text{max}}} = \frac{-\sum_{i=1}^{n} p_{u,i} \log_b p_{u,i}}{\log_b n}, 
\]
where $H(P_u) = -\sum_{i=1}^{n} p_{u,i} \log_b p_{u,i}$ is the shannon entropy of the distribution and $H_{\text{max}} = \log_b n$ is the maximum possible entropy occurs when the distribution is uniform. 

The range of $Div(P)$ is in $[0, 1]$, where a low values means the predictions are concentrated in a few classes, i.e., less diversity, and a high value means the predictions are spread across many classes, i.e., high diversity. 

For text generation tasks, we cannot directly apply this metric since the output space is continuous. To adapt it, we first convert the generated texts into embeddings and then apply a clustering algorithm (e.g., \( k \)-means) to discretize the continuous space. This yields a discrete distribution over clusters for each user. Consequently, we obtain cluster-based distributions for the generated texts, analogous to class distributions in classification tasks, which can then be used to compute diversity via normalized entropy.

\section{Black-Box Local-Global Memory Modeling}

Our LoGo framework can also be implemented for black-box LLMs where the model parameters are not accessible. For such methods, we use prompts to inject the additional knowledge into the model, guiding the model's behavior instead of updating model parameters. 

\subsection{Modeling Evolving Global Memory}
\label{sec:global-memory-blackbox}

To enable adaptation to shifting user behavior and population-wide trends, it is critical to maintain a global memory that evolves over time. In black-box settings, where the underlying model parameters cannot be directly accessed or fine-tuned, we instead represent global memory using external artifacts, such as textual summaries or retrieved exemplars.

\input{table/non-parametric.tex}

The process for constructing and updating global memory proceeds in structured time phases. Specifically, we partition the complete user interaction history into \( T \) non-overlapping temporal segments $\{\mathcal{H}^{(1)}, \mathcal{H}^{(2)}, \dots, \mathcal{H}^{(T)}\}$ where each \( \mathcal{H}^{(t)} \) contains all user interactions from time phase \( t \). At each phase \( t \), the update procedure involves two key steps:

\begin{enumerate}
    \item \textbf{User Profile Update:} For each user, we generate a concise summary of their behavior within \( \mathcal{H}^{(t)} \). These profiles may consist of selected queries, preferences, or representative interactions, formatted as textual snippets or lightweight exemplars. The profiles are designed to be compatible with black-box LLMs, serving as conditioning context or part of the prompt. We take the movie tagging task as an example: 
    \begin{quote}
    \textit{Prompt:} ``You will receive a personalized memory between the movies and their taggings along with a list of new movies and their descriptive tags. Update the memory by identifying the correlations between the movies and their tags, while preserving key insights from the original. 
    Personalized memory: \texttt{[updated personalized memory]}.
    New movies with tags: \texttt{[A list of movies with tags]}''
    \end{quote}

    \item \textbf{Global Memory Update:} The individual user profiles from phase \( t \) are aggregated into a population-level summary, which captures dominant trends and shared behaviors during that time period. This summary is appended to or used to revise the evolving global memory \( \mathcal{M}_G \). The update is cumulative, preserving past insights while allowing temporal adaptation. Also, we take the movie tagging task as an example: 
    \begin{quote}
    \textit{Prompt:} ``You will receive a global memory and a set of personalized memories. Update the global memory by identifying patterns between paper titles and their most relevant references, while preserving key insights from the original. Output a bulleted list with 20 items. 
    Global memory: \texttt{[updated global memory]}.
    Personalized memories: \texttt{[A list of personalized memories]}.''
    \end{quote}

\end{enumerate}

This time-aware aggregation ensures that the global memory reflects the evolution of population behavior without requiring access to or modification of the underlying model. The final memory \( \mathcal{M}_G \), constructed over all \( T \) phases, serves as an external, temporally-aware context to guide personalized inference or recommendation tasks.

\subsection{Mitigating Conflicts Between Local and Global Memory}
\label{sec:local-global-blackbox}

The combination of local memory \( \mathcal{M}_L \) and global memory \( \mathcal{M}_G \) enables black-box LLMs to balance personalization with generalization. However, these memory sources may occasionally provide conflicting signals, particularly when an individual user's preferences diverge from dominant population trends. A central challenge is to determine whether such divergence reflects persistent user behavior or a transient anomaly. Resolving this is critical to appropriately weighting local versus global memory in prediction.

In black-box settings, where model parameters are inaccessible, we resolve this conflict via a prompt-based \emph{mediator}. Instead of modifying the model or training a parameterized adapter, we instantiate the mediator as a structured prompt template that combines local and global memory representations into a single coherent context.

\noindent\textbf{Local Memory Construction.}
The local memory \( \mathcal{M}_L \) is constructed from the user's historical interactions \( \mathcal{H}_u^{<t} \) using one of the following methods:
\begin{enumerate}
    \item \textit{RAG:} Retrieve the top-\( k \) past interactions most relevant to the current query \( q \):
    \[
    \mathcal{M}_L = \mathcal{R}(q, \mathcal{H}_u^{<t}, k)
    \]
    \item \textit{Profile:} Generate a user profile summary from the interaction history:
    \[
    p = \text{LLM}(\mathcal{H}_u^{<t}), \quad \mathcal{M}_L = p
    \]
    \item \textit{Hybrid Memory:} Combine retrieved examples with the profile:
    \[
    \mathcal{M}_L = \mathcal{R}(q, \mathcal{H}_u^{<t}, k) \cup p
    \]
\end{enumerate}

\noindent\textbf{Global Memory Access.}
The global memory \( \mathcal{M}_G \) is provided as an externally constructed summary of population-level behavior, obtained through the evolving temporal aggregation process described in Section~\ref{sec:global-memory-blackbox}. It is formatted as a natural language summary or exemplar list and passed into the prompt.

\noindent\textbf{Prompt-Based Mediator.}
We define a \textbf{mediator prompt} \( \mathcal{F}_{\text{prompt}} \) that adaptively integrates local and global memory. We take the movie tagging task as an example:
\begin{quote}
    \textit{Prompt:}``Here is the current user's memory: \texttt{[local memory]}. Here is the global memory: \texttt{[global memory]}. You need to balance their contributions. \\
    Which tag does this movie relate to among the following tags? Just answer with the tag name without further explanation. tags: [...]
    description: \texttt{[movie descriptions]} tag:''
\end{quote}
This prompt serves as a soft mediator, implicitly resolving conflicts by allowing the LLM to condition on both memory sources within a unified context. Because the mediation logic is encoded at the prompt level, no fine-tuning or architecture modification is required.

\noindent\textbf{Inference.}
Given a query \( q_u \), the black-box prediction is:
\[
\mathcal{F}_u^*(q_u) = \text{LLM}(\mathcal{F}_{\text{prompt}}(q_u, \mathcal{M}_L, \mathcal{M}_G))
\]
This formulation preserves the generalization encoded in \( \mathcal{M}_G \), while allowing user-specific signals from \( \mathcal{M}_L \) to influence the response. Different strategies for memory formatting (e.g., interleaved examples, bullet summaries) can be used depending on the application and prompt length constraints. This prompt-based architecture enables efficient adaptation, extensibility to new users, and compatibility with API-only LLM deployments.

\subsection{Black-Box Results}

We evaluate the effectiveness of LoGo in a black-box setting, where both global and local memory are modeled externally via prompts without access to model parameters. The results are presented in Table~\ref{tab:non-parametric}, and several key observations emerge:

\begin{itemize}
    \item \textbf{Personalization Improves Performance.} Similar to the white-box case, the use of local memory, either via RAG or profile summaries, consistently outperforms the non-personalized base model and random baselines. This confirms that even without parameter tuning, black-box LLMs benefit from personalization through prompt augmentation.
    
    \item \textbf{Global Memory Offers Complementary Gains.} Incorporating global memory via LoGo yields additional improvements. Specifically, LoGo variants that include global summaries outperform their base model counterparts across all metrics, indicating the value of shared population-level knowledge in supporting generalization.
    
    \item \textbf{Mediator Prompt Effectively Resolves Conflicts.} Fusing local and global memory through a structured mediator prompt leads to the best results. The \textit{LoGo + Profile w. RAG} variant achieves the strongest performance across nearly all benchmarks, outperforming other methods in both accuracy and generation quality (e.g., F1, ROUGE).
    
    \item \textbf{White-Box vs. Black-Box Trade-Off.} While the black-box setting yields slightly lower overall performance than the white-box setting, the gap remains modest. This demonstrates that prompt-based memory integration is a practical and effective strategy when direct model access is unavailable.
\end{itemize}

Overall, these findings demonstrate the robustness of LoGo across different access regimes. Even under strict black-box constraints, LoGo delivers strong performance by leveraging personalized local signals and population-level trends, integrated via a prompt-based mediator.

\end{document}

%% file: math_commands.tex
\usepackage{amsmath,amsfonts,bm}

\def\eqref#1{equation~\ref{#1}}

\def\1{\bm{1}}

\DeclareMathAlphabet{\mathsfit}{\encodingdefault}{\sfdefault}{m}{sl}
\SetMathAlphabet{\mathsfit}{bold}{\encodingdefault}{\sfdefault}{bx}{n}

\DeclareMathOperator*{\argmin}{arg\,min}

%% file: method.tex
\section{LoGo for LLM Personalization}

LLM Personalization typically involves using a user’s past interactions to tailor the model in order to infer future user behaviors. Given a user \( u \in \mathcal{U} \), the goal is to generate a personalized output \( r \) conditioned on the current input \( q \) and the user's historical behavior \( \mathcal{H}_u = \{h_i\}_{i=1}^{n} \), where each record \( h_i \) corresponds to a query–response pair \( (q, r, t) \) at timestamp \( t \).  
In this paper, we propose LoGo to achieve personalization, as shown in Figure \ref{fig:framework}. LoGo relying on two complementary memory components:  
\textit{local memory}, which captures fine-grained preferences unique to an individual user, and \textit{global memory}, which encodes collective behavioral patterns shared across the population.   
The local memory is defined as
$\mathcal{M}_L = \mathcal{R}(\mathcal{H}_u^{<t})$,
where \( \mathcal{R}(\cdot) \) is a function that extracts relevant information from the past interactions of user \( u \) prior to time \( t \).  
The global memory is defined as
$
\mathcal{M}_G = \mathcal{S}(\{\mathcal{H}_v\}_{v \in \mathcal{U}})$,
where \( \mathcal{S}(\cdot) \) aggregates the histories of all users \( v \in \mathcal{U} \) to distill population-level trends.  We defer the detailed construction of \( \mathcal{M}_L \) and \( \mathcal{M}_G \) to subsequent sections. Our goal is to learn user-specific parameters \( \Theta_u^* \) to minimize the task loss:
\begin{equation}
    \Theta_u^* = \arg\min_{\Theta} \, \sum_{(q, r, t) \sim \mathcal{H}_u} 
    \mathcal{L}(q, r; \Theta, \mathcal{M}_L, \mathcal{M}_G),
    \label{eq:obj}
\end{equation}
where \( \mathcal{L} \) is the task-specific loss function, which for text generation corresponds to the next-token prediction loss. However, effectively realizing this formulation in practice requires addressing two key challenges:




\begin{itemize}
    \item \textbf{Modeling evolving global memory:}  User interests shift over time \citep{wu2017modeling}, so global memory at time \( T \) should incorporate all prior steps while weighing recent signals more heavily. Persistent patterns from earlier phases are valuable (e.g., time step \( T{-}1 \)), but older information  should gradually lose influence.  
    
    \item \textbf{Resolving local-global memory conflicts:} Local and global memory capture individual and collective interests, but may sometimes diverge \citep{zheng2021disentangling}. For example, a user might prefer niche films while the broader population tends to favor mainstream content. The key challenge is to distinguish persistent from situational preferences and adjust memory influence to improve prediction accuracy. 

\end{itemize}
We address these challenges in the following sections.

\subsection{Modeling Evolving Global Memory}
\label{sec:global-memory}

To support generalization to future behavior, the global memory \( \mathcal{M}_G \) must capture population-level trends that evolve over time, rather than relying on static aggregates. In real-world systems, user interests and collective patterns shift due to external events, seasonal cycles, or cultural dynamics \citep{zheng2021disentangling}. Encoding outdated information can degrade personalization quality. For example, in news recommendation systems, attention to specific topics may surge and dissipate rapidly, making it essential to model temporal dynamics \citep{shen2022session}.

To capture these evolving patterns, we model \( \mathcal{M}_G \) as a parameterized module that evolves through a time-aware training strategy. Specifically, we partition the full set of user interaction histories \( \{\mathcal{H}_v\}_{v \in \mathcal{U}} \) into \( T \) non-overlapping temporal segments $\{\mathcal{H}_v\}_{v \in \mathcal{U}} = \{\mathcal{H}^{(1)}, \mathcal{H}^{(2)}, \dots, \mathcal{H}^{(T)}\}$ where each \( \mathcal{H}^{(t)} \) contains interactions from all users during time phase \( t \). At each phase, the global memory is parameterized by \( \Phi_{G,t} \) and optimized to minimize the task loss on the current segment \( \mathcal{H}^{(t)} \):
\begin{equation}
    \Phi_{G,t}^* = \argmin_{\Phi_{G,t} \xleftarrow{} \Phi_{G,t-1}^*} \sum_{(q, r) \in \mathcal{H}^{(t)}} \mathcal{L}(q, r; \Theta, \Phi_{G,t}),
    \label{eq:global}
    \end{equation}
where \( \Theta \) denotes the frozen base model parameters and \( \Phi_{G,t} \) are the trainable adapter parameters of the global memory module at time phase \( t \).

In Equation~\ref{eq:global}, the optimization parameters $\Phi_{G,t}$ are warm-started from the previous optimal parameters \( \Phi_{G,t-1}^* \). This evolving procedure is crucial: (1) It preserves historical knowledge while gradually down-weighting outdated patterns \citep{hardt2013decay}. In effect, each \( \Phi_{G,t} \) encodes a temporally smoothed representation of population behavior, and can be interpreted as a realization of the evolving trend. (2) It allows the model to adaptively reweight the influence of different time periods without requiring manually designed decay schedules or heuristic weighting. (3) It is substantially more efficient than retraining from scratch at each phase, as population-level trends typically shift incrementally rather than abruptly. The final global memory used for personalization, denoted by \( \mathcal{M}_G \), is characterized by \( \Phi_{G,T}^* \). It represents the endpoint of a temporally evolving process that summarizes the progression of population behavior over time.

Importantly, the parameterized global memory \( \Phi_{G,t} \) is implementation agnostic. It can be instantiated using lightweight modules such as LoRA~\cite{hu2021lora} or prefix tuning~\cite{li-liang-2021-prefix}, or replaced with more expressive memory architectures depending on system constraints and resource availability. In black-box settings where the base model \( \Theta \) is inaccessible, the memory can instead be represented as retrieved exemplars or textual summaries. This flexibility makes the framework broadly applicable across a range of deployment scenarios. While Equation~\ref{eq:global} addresses the white-box setting, we defer the black-box variant to the appendix.

\input{table/parametric.tex}

\subsection{Mitigating Conflicts Between Local and Global Memory}

While the combination of local memory \( \mathcal{M}_L \) and global memory \( \mathcal{M}_G \) allows the model to balance personalization and generalization, these two sources can sometimes produce conflicting signals. Such conflicts typically arise when a user's preferences diverge from population-level behavior. A key challenge in resolving this conflict is to determine whether the divergence reflects a persistent personal trait or a collective deviation. This distinction is critical because it informs how much influence each memory component should have. If a user consistently prefers niche content, local memory should be prioritized; if the preference is transient, global memory may offer a more reliable prior. 


To address this, we introduce a mediator module \( \mathcal{F} \), characterized by parameters \( \Phi_L \), which adaptively integrates local and global memory representations. The mediator is adaptable to different architectures and resource constraints.  
In the white-box setting, the mediator is either jointly trained with the base model or added as a lightweight module (e.g., adapters) to modulate local and global signals. In the black-box setting, it is approximated via prompts that encode memory externally. This section focuses on the white-box case; the black-box variant is discussed in the appendix.

Before introducing the mediator, we first define the local memory \( \mathcal{M}_L \).  
 Following~\citet{tan2024oppu},  we define \( \mathcal{M}_L \) as a query-dependent representation constructed from the user’s historical interactions \( \mathcal{H}_u^{<t} \), transformed by a function \( \phi \). We consider two variants of \( \phi \):  
(i) \(\phi_{\text{r}}\), which employs \textit{Retrieval-Augmented Generation (RAG)}~\cite{lewis2020retrieval} to retrieve the top-\( k \) most relevant past interactions for the current query \( q \), i.e., $\mathcal{M}_L = \mathcal{R}(q, \mathcal{H}_u^{<t}, k)$;
(ii) \(\phi_{\text{p}}\), which extends RAG through \textit{Profile-Augmented Generation (PAG)}~\cite{richardson2023integrating} by summarizing a user profile $p = \text{LLM}(\mathcal{H}_u^{<t})$.

Next we train a personalized mediator module \( \mathcal{F} \), parameterized by \( \Phi_L \), for each user. The mediator is optimized using supervised query-response pairs sampled from the user’s interaction history:
\begin{equation}
    \Phi_L^* = \arg\min_{\Phi_L} 
    \sum_{(q, r, t) \in \mathcal{H}_u} 
    \mathcal{L}(\tilde{q}, r; \Theta, \mathcal{M}_G, \Phi_L),
\end{equation}
where \( \tilde{q} = \phi(q, \mathcal{M}_L) \). 
During training, only the mediator is updated, while the base model \( \Theta \), the local memory \( \mathcal{M}_L \), and the global memory \( \mathcal{M}_G \) (parameterized by \( \Phi_{G,T}^* \)) remain frozen. This preserves the generalization capability of the global memory and the personalized signals encoded in the local memory, while enabling efficient user-specific adaptation through the mediator. 
Finally, during inference, given a user query \( q_u \), the model prediction is:
\[
\mathcal{F}_{u}^*(q_u) =
\mathcal{F}_{\text{med}} \big(
\mathcal{F}_{\Theta, \mathcal{M}_G}(\tilde{q}_u)\big),
\]
where \( \tilde{q}_u = \phi(q_u, \mathcal{M}_L) \) is the query augmented with local memory,
\( \mathcal{F}_{\Theta, \mathcal{M}_G} \) denotes the base model \(\Theta\) integrated with global memory \(\mathcal{M}_G\), and
\( \mathcal{F}_{\text{med}} \) adaptively fuses local and global signals to generate the final prediction.

\subsection{Community-Aware Global Memory}

While global memory modeled at the population level enables broad generalization, it risks diluting subgroup-specific patterns, particularly in domains with highly diverse user interests. For instance, preferences may differ significantly between distinct communities such as professionals, students, or fans of particular genres. Treating all users as a single homogeneous distribution may introduce noise and reduce the strength of representation of personalization. To address this limitation, we extend our framework by introducing \textit{community memory}, a collection of global memory modules, each specialized for a particular user community. This design captures shared knowledge at a finer granularity and enhances alignment with the behavioral patterns of distinct subgroups.

We begin by deriving a profile vector \( \boldsymbol{\rho}_u \) for each user, summarizing their historical interactions. Each profile vector is computed as the mean of the embedding representations of the user's past queries and responses: 
\[
\boldsymbol{\rho}_u = \frac{1}{|\mathcal{H}_u|} 
\sum\nolimits_{(q, r) \in \mathcal{H}_u} 
\big( \operatorname{emb}(q) \oplus \operatorname{emb}(r) \big)
\]
where \( \oplus \) denotes vector concatenation. Collecting all user profiles yields the set \( \mathcal{P} = \{\boldsymbol{\rho}_1, \dots, \boldsymbol{\rho}_m\} \). We then apply \( k \)-means clustering over \( \mathcal{P} \) to partition users into \( K \) distinct communities \( \{ \mathcal{C}_1, \dots, \mathcal{C}_K \} \). 
For each community \( \mathcal{C}_k \), we construct a dedicated global memory module \( \Phi_{G,t}^k \) by training on interaction data restricted to that community. Compared to the single global memory \( \Phi_{G,t} \) trained over all users, community-specific modules capture more nuanced and context-specific patterns, better reflecting the shared preferences and behavioral trends within each subgroup. 

%% file: table/parametric.tex
\begin{table*}[!t]
    \centering
    \resizebox{\linewidth}{!}{
        \begin{tabular}{lcccccccccc}
            \toprule
            \multirow{2}{*}{\bf Method}                   & \multicolumn{2}{c}{\textbf{\textsc{LaMP 2N}}} & \multicolumn{2}{c}{\textbf{\textsc{LaMP 2M}}} & \multicolumn{2}{c}{\textbf{\textsc{LaMP 3}}} & \multicolumn{2}{c}{\textbf{\textsc{LaMP 4}}} & \multicolumn{2}{c}{\textbf{\textsc{LaMP 5}}}                                                                                                                                                                                               \\ \cmidrule(lr){2-3} \cmidrule(lr){4-5} \cmidrule(lr){6-7} \cmidrule(lr){8-9} \cmidrule(lr){10-11}
                                                          & Acc $\uparrow$                                & F1  $\uparrow$                                & Acc $\uparrow$                               & F1  $\uparrow$                               & MAE $\downarrow$                             & RMSE $\downarrow$                         & R-1 $\uparrow$ & R-L $\uparrow$                           & R-1 $\uparrow$                           & R-L $\uparrow$                           \\

            \midrule
            Base Model                                    & 0.791                                         & 0.536                                         & 0.542                                        & 0.503                                        & 0.277                                        & 0.295                                     & 0.203          & 0.183                                    & 0.522                                    & 0.457                                    \\
            Base Model + Rand.                            & 0.802                                         & 0.577                                         & 0.569                                        & 0.506                                        & 0.232                                        & 0.281                                     & 0.208          & 0.186                                    & 0.520                                    & 0.465                                    \\

            OPPU                                          & 0.807                                         & 0.606                                         & 0.702                                        & 0.612                                        & 0.218                                        & 0.218                                     & 0.216          & 0.195                                    & 0.519                                    & 0.465                                    \\
            OPPU + RAG                                    & 0.817                                         & 0.580                                         & 0.588                                        & 0.524                                        & 0.223                                        & 0.259                                     & 0.205          & 0.185                                    & 0.521                                    & 0.468                                    \\
            OPPU + PAG                                    & 0.814                                         & 0.608                                         & 0.581                                        & 0.528                                        & 0.250                                        & 0.339                                     & 0.208          & 0.187                                    & 0.513                                    & 0.453                                    \\

            \midrule
            HYDRA                                         & 0.780                                         & 0.401                                         & 0.540                                        & 0.458                                        & 0.400                                        & 0.747                                     & 0.178          & 0.169                                    & 0.434                                    & 0.372                                    \\
            PER-PCS                                       & 0.804                                         & 0.539                                         & 0.679                                        & 0.583                                        & 0.251                                        & 0.244                                     & 0.205          & 0.185                                    & 0.520                                    & 0.471                                    \\

            \midrule
            \rowcolor{Gray}           LoGo (c = 1)        & 0.824                                         & 0.611                                         & 0.710                                        & 0.630                                        & 0.188                                        & 0.205                                     & \textbf{0.216} & \textbf{0.196}                           & 0.516                                    & 0.470                                    \\
            \rowcolor{Gray}           LoGo + RAG (c = 1)  & 0.828                                         & 0.614                                         & 0.724                                        & 0.642                                        & \textbf{0.161}                               & \textbf{0.196}                            & 0.214          & 0.194                                    & 0.526                                    & 0.477                                    \\
            \rowcolor{Gray}           LoGo + PAG (c = 1)  & 0.842                                         & 0.627                                         & 0.724                                        & 0.644                                        & 0.196                                        & 0.268                                     & 0.213          & 0.193                                    & 0.527                                    & 0.479                                    \\

            \midrule
            \rowcolor{Gray}           LoGo + PAG (c = 5)  & \textbf{0.854}                                & \textbf{0.650}                                & 0.739                                        & 0.668                                        & 0.186                                        & 0.238                                     & 0.212          & 0.193                                    & 0.536                                    & 0.481                                    \\
            \rowcolor{Gray}           LoGo + PAG (c = 10) & 0.852                                         & 0.649                                         & \textbf{0.741}                               & \textbf{0.689}                               & 0.193                                        & 0.201                                     & 0.199          & 0.184                                    & \textbf{0.538}                           & \textbf{0.483}                           \\
            \rowcolor{Gray}           LoGo + PAG (c = 20) & 0.849                                         & 0.640                                         & 0.740                                        & 0.673                                        & 0.248                                        & 0.237                                     & 0.211          & 0.192                                    & 0.516                                    & 0.472                                    \\

            \midrule
            \rowcolor{Gray} \textit{Improvement}          & \textcolor{DarkGreen}{4.53\% $\uparrow$}      & \textcolor{DarkGreen}{6.91\% $\uparrow$}      & \textcolor{DarkGreen}{5.56\% $\uparrow$}     & \textcolor{DarkGreen}{12.58\% $\uparrow$}    & \textcolor{DarkGreen}{26.15\% $\uparrow$}    & \textcolor{DarkGreen}{10.09\% $\uparrow$} & 0.00\%         & \textcolor{DarkGreen}{0.51\% $\uparrow$} & \textcolor{DarkGreen}{3.07\% $\uparrow$} & \textcolor{DarkGreen}{3.21\% $\uparrow$} \\
            \bottomrule
        \end{tabular}
    }
    \caption{
        Experimental results of the white-box implementation of LoGo using LLaMA 3.1–8B. R-1 and R-L denote ROUGE-1 and ROUGE-L, respectively. $\uparrow$ indicates that higher values are better, while $\downarrow$ indicates that lower values are preferred. \textbf{Bold} indicates the best results and \textcolor{DarkGreen}{green} shows the improvements of the best results over the best baselines.
    }
    \label{tab:parametric}
\end{table*}

%% file: table/cold-start.tex
\begin{table}[!t]
    \resizebox{\linewidth}{!}{
        \begin{tabular}{lccccc}
            \toprule
            \multirow{2}{*}{Method} & \multirow{2}{*}{Variant} & \multicolumn{2}{c}{\textbf{\textsc{LaMP-2N}}} & \multicolumn{2}{c}{\textbf{\textsc{LaMP-4}}}                                                                                       \\ \cmidrule(lr){3-4} \cmidrule(lr){5-6}
                                    &                          & Acc $\uparrow$                                & F1  $\uparrow$                               & R-1 $\uparrow$                           & R-L $\uparrow$                           \\
            \midrule
            LoGo                    & w.o. $\mathcal{M}_G$     & 0.620                                         & 0.526                                        & 0.188                                    & 0.170                                    \\
            -Base                   & w. $\mathcal{M}_G$       & 0.667                                         & 0.579                                        & 0.200                                    & 0.182                                    \\ \cmidrule(lr){2-6}
                                    & \textit{Imp.}            & \textcolor{DarkGreen}{7.58\% $\uparrow$}      & \textcolor{DarkGreen}{10.08\% $\uparrow$}    & \textcolor{DarkGreen}{6.38\% $\uparrow$} & \textcolor{DarkGreen}{7.06\% $\uparrow$} \\
            \midrule
            LoGo                    & w.o. $\mathcal{M}_G$     & 0.576                                         & 0.493                                        & 0.185                                    & 0.166                                    \\
            -RAG                    & w. $\mathcal{M}_G$       & 0.656                                         & 0.582                                        & 0.190                                    & 0.171                                    \\ \cmidrule(lr){2-6}
                                    & \textit{Imp.}            & \textcolor{DarkGreen}{13.89\% $\uparrow$}     & \textcolor{DarkGreen}{18.05\% $\uparrow$}    & \textcolor{DarkGreen}{2.70\% $\uparrow$}  & \textcolor{DarkGreen}{3.01\% $\uparrow$} \\
            \bottomrule
        \end{tabular}
    }
    \caption{
        Performance comparison on inactive (cold-start) users, defined as the bottom 25\% of users with the fewest historical interactions. The results demonstrate that incorporating global memory $\mathcal{M}_G$ significantly enhances performance by leveraging collective behavioral patterns to guide individual user predictions.
    }
    \label{tab:cold-start}
\end{table}

%% file: table/biasing.tex
\begin{table}[!t]
    \resizebox{\linewidth}{!}{
        \begin{tabular}{lccccc}
            \toprule
            \multirow{2}{*}{Method} & \multirow{2}{*}{Variant} & \multicolumn{2}{c}{\textbf{\textsc{LaMP-2M}}} & \multicolumn{2}{c}{\textbf{\textsc{LaMP-4}}}                                                                                       \\ \cmidrule(lr){3-4} \cmidrule(lr){5-6}
                                    &                          & F1 $\uparrow$                                 & Div. $\uparrow$                              & R-L $\uparrow$                           & Div. $\uparrow$                          \\
            \midrule
            LoGo                    & w.o. $\mathcal{M}_G$     & 0.509                                         & 0.703                                        & 0.181                                    & 0.953                                    \\
            -Base                   & w. $\mathcal{M}_G$       & 0.604                                         & 0.736                                        & 0.195                                    & 0.962                                    \\ \cmidrule(lr){2-6}
                                    & \textit{Imp.}            & \textcolor{DarkGreen}{18.66\% $\uparrow$}     & \textcolor{DarkGreen}{4.69\% $\uparrow$}     & \textcolor{DarkGreen}{7.73\% $\uparrow$} & \textcolor{DarkGreen}{0.94\% $\uparrow$} \\
            \midrule
            LoGo                    & w.o. $\mathcal{M}_G$     & 0.453                                         & 0.745                                        & 0.183                                    & 0.951                                    \\
            -RAG                    & w. $\mathcal{M}_G$       & 0.506                                         & 0.780                                        & 0.186                                    & 0.965                                    \\ \cmidrule(lr){2-6}
                                    & \textit{Imp.}            & \textcolor{DarkGreen}{11.70\% $\uparrow$}      & \textcolor{DarkGreen}{4.70\% $\uparrow$}      & \textcolor{DarkGreen}{1.64\% $\uparrow$} & \textcolor{DarkGreen}{1.47\% $\uparrow$} \\
            \bottomrule
        \end{tabular}
    }
    \caption{
        Performance comparison on highly active users, defined as the top 25\% of users with the most historical interactions. The results show that integrating global memory $\mathcal{M}_G$ consistently improves both performance and diversity (Div.), highlighting its effectiveness in mitigating user-level biases by leveraging shared behavioral patterns.
    }
    \label{tab:biasing}
\end{table}

%% file: table/ablation.tex
\begin{table*}[!t]
    \resizebox{\linewidth}{!}{
        \begin{tabular}{llcccccccccc}
            \toprule
            \multirow{2}{*}{\bf Setting} & \multirow{2}{*}{\bf Variant} & \multicolumn{2}{c}{\textbf{\textsc{LaMP-2N}}} & \multicolumn{2}{c}{\textbf{\textsc{LaMP-2M}}} & \multicolumn{2}{c}{\textbf{\textsc{LaMP-3}}} & \multicolumn{2}{c}{\textbf{\textsc{LaMP-4}}} & \multicolumn{2}{c}{\textbf{\textsc{LaMP-5}}}                                                                                         \\ \cmidrule(lr){3-4} \cmidrule(lr){5-6} \cmidrule(lr){7-8} \cmidrule(lr){9-10} \cmidrule(lr){11-12}
                                         &                              & Acc $\uparrow$                                & F1 $\uparrow$                                 & Acc $\uparrow$                               & F1 $\uparrow$                                & MAE $\downarrow$                             & RMSE $\downarrow$ & R-1 $\uparrow$ & R-L $\uparrow$ & R-1 $\uparrow$ & R-L $\uparrow$ \\ \midrule
                                         & T = 1                        & 0.807                                         & 0.602                                         & 0.702                                        & 0.619                                        & 0.297                                        & 0.274             & 0.201          & 0.182          & 0.504          & 0.465          \\
            \textbf{Time Splits}         & T = 5                        & 0.842                                         & 0.627                                         & 0.724                                        & 0.644                                        & \textbf{0.196}                               & \textbf{0.268}    & 0.213          & 0.193          & \textbf{0.527} & \textbf{0.479} \\
                                         & T = 10                       & \textbf{0.859}                                & \textbf{0.638}                                & \textbf{0.737}                               & \textbf{0.656}                               & 0.204                                        & 0.285             & \textbf{0.214} & \textbf{0.196} & 0.524          & 0.478          \\
            \midrule
                                         & k = 1                        & 0.842                                         & 0.627                                         & 0.724                                        & 0.644                                        & 0.196                                        & 0.268             & 0.213          & 0.193          & 0.527          & 0.479          \\
            \textbf{Retrieved Items}     & k = 2                        & 0.850                                         & 0.636                                         & 0.727                                        & 0.651                                        & \textbf{0.192}                               & \textbf{0.265}    & \textbf{0.221} & \textbf{0.197} & 0.537          & 0.484          \\
                                         & k = 4                        & \textbf{0.856}                                & \textbf{0.642}                                & \textbf{0.733}                               & \textbf{0.654}                               & 0.207                                        & 0.281             & 0.206          & 0.189          & \textbf{0.541} & \textbf{0.488} \\

            \bottomrule
        \end{tabular}
    }
    \caption{
    Hyper-parameter analysis on key design choices in the LoGo. We evaluate (1) the number of temporal splits used to update global memory (default T=5), and (2) the number of retrieved history items (default k=1). 
    }
    \label{tab:ablation}
\end{table*}

%% file: table/non-parametric.tex
\begin{table*}[!t]
    \centering
    \resizebox{\linewidth}{!}{
        \begin{tabular}{lcccccccccc}
            \toprule
            \multirow{2}{*}{\bf Method}                                        & \multicolumn{2}{c}{\textbf{\textsc{LaMP-2N}}} & \multicolumn{2}{c}{\textbf{\textsc{LaMP-2M}}} & \multicolumn{2}{c}{\textbf{\textsc{LaMP-3}}} & \multicolumn{2}{c}{\textbf{\textsc{LaMP-4}}} & \multicolumn{2}{c}{\textbf{\textsc{LaMP-5}}}                                                                                         \\ \cmidrule(lr){2-3} \cmidrule(lr){4-5} \cmidrule(lr){6-7} \cmidrule(lr){8-9} \cmidrule(lr){10-11}
                                                                               & Acc $\uparrow$                                & F1 $\uparrow$                                 & Acc $\uparrow$                               & F1 $\uparrow$                                & MAE $\downarrow$                             & RMSE $\downarrow$ & R-1 $\uparrow$ & R-L $\uparrow$ & R-1 $\uparrow$ & R-L $\uparrow$ \\
            \midrule
            Base Model                                                         & 0.744                                         & 0.520                                         & 0.457                                        & 0.379                                        & 0.304                                        & 0.582             & 0.159          & 0.142          & 0.508          & 0.417          \\
            Base Model + Rand.                                                 & 0.755                                         & 0.536                                         & 0.513                                        & 0.398                                        & 0.318                                        & 0.563             & 0.159          & 0.156          & 0.423          & 0.355          \\
            \midrule
            Base Model + RAG                                                   & 0.762                                         & 0.553                                         & 0.513                                        & 0.420                                        & 0.259                                        & 0.543             & 0.177          & 0.159          & 0.507          & 0.427          \\
            Base Model + Profile                                               & 0.798                                         & 0.608                                         & 0.556                                        & 0.454                                        & {0.241}                                      & 0.538             & 0.162          & 0.145          & 0.529          & {0.435}        \\
            Base Model + Profile w. RAG                                        & 0.809                                         & 0.621                                         & 0.581                                        & 0.472                                        & 0.250                                        & {0.535}           & {0.193}        & {0.173}        & 0.512          & 0.432          \\
            \midrule
            \rowcolor{Gray}                              LoGo                  & 0.757                                         & 0.537                                         & 0.504                                        & 0.409                                        & 0.290                                        & 0.562             & 0.186          & 0.167          & 0.524          & 0.421          \\
            \rowcolor{Gray}                              LoGo + RAG            & 0.774                                         & 0.579                                         & 0.524                                        & 0.432                                        & 0.250                                        & 0.535             & 0.191          & 0.171          & 0.515          & 0.434          \\
            \rowcolor{Gray}                              LoGo + Profile        & {0.822}                                       & {0.652}                                       & 0.589                                        & {0.484}                                      & 0.286                                        & 0.567             & 0.182          & 0.163          & \textbf{0.534} & \textbf{0.439} \\
            \rowcolor{Gray}                              LoGo + Profile w. RAG & \textbf{0.828}                                & \textbf{0.658}                                & \textbf{0.601}                               & \textbf{0.498}                               & \textbf{0.239}                               & \textbf{0.523}    & \textbf{0.203} & \textbf{0.182} & {0.532}        & 0.427          \\
            \bottomrule
        \end{tabular}
    }
    \caption{
        Experimental results of the white-box implementation of LoGo using Claude 3.7. R-1 and R-L denote ROUGE-1 and ROUGE-L, respectively. $\uparrow$ indicates that higher values are better, while $\downarrow$ indicates that lower values are preferred. \textbf{Bold} indicates the best results.
    }
    \label{tab:non-parametric}
\end{table*}